\documentclass[conference]{IEEEtran}
\IEEEoverridecommandlockouts
\usepackage{cite}
\usepackage{tabularx}
\usepackage{caption} 
\usepackage{amsmath,amssymb,amsfonts}
\usepackage{graphicx}
\usepackage{multirow}
\usepackage{float}
\usepackage{textcomp}
\usepackage{xcolor}
\usepackage[symbol]{footmisc}

\def\BibTeX{{\rm B\kern-.05em{\sc i\kern-.025em b}\kern-.08em
    T\kern-.1667em\lower.7ex\hbox{E}\kern-.125emX}}
\begin{document}

\title{Evolution of Transfer Learning in Natural Language Processing\\
}

\author{
\IEEEauthorblockN{Aditya Malte \IEEEauthorrefmark{2}}
\IEEEauthorblockA{{Dept. of Computer Engineering,} \\
Pune Institute of Computer Technology,\\
Maharashtra, India.}
\and
\IEEEauthorblockN{Pratik Ratadiya \IEEEauthorrefmark{2}}
\IEEEauthorblockA{{Dept. of Computer Engineering,} \\
Pune Institute of Computer Technology,\\
Maharashtra, India.}
}

\maketitle
\begin{abstract}
In this paper, we present a study of the recent advancements which have helped bring Transfer Learning to NLP through the use of semi-supervised training. We discuss cutting-edge methods and architectures such as BERT, GPT, ELMo, ULMFit among others. Classically, tasks in natural language processing have been performed through rule-based and statistical methodologies. However, owing to the vast nature of natural languages these methods do not generalise well and failed to learn the nuances of language. Thus machine learning algorithms such as Naive Bayes and decision trees coupled with traditional models such as Bag-of-Words and N-grams were used to usurp this problem. Eventually, with the advent of advanced recurrent neural network architectures such as the LSTM, we were able to achieve state-of-the-art performance in several natural language processing tasks such as text classification and machine translation.  We talk about how Transfer Learning has brought about the well-known ImageNet moment for NLP. Several advanced architectures such as the Transformer and its variants have allowed practitioners to leverage knowledge gained from unrelated task to drastically fasten convergence and provide better performance on the target task. This survey represents an effort at providing a succinct yet complete understanding of the recent advances in natural language processing using deep learning in with a special focus on detailing transfer learning and its potential advantages.
\end{abstract}

\begin{IEEEkeywords}
Natural language processing, transfer learning, self attention, language modeling
\end{IEEEkeywords}

\section{INTRODUCTION}
Natural Language processing - the science of how to make computers effectively process natural text- has recently witnessed rapid advancements thanks to increased processing power, data and better algorithms. It forms the heart of several use cases such as opinion mining, conversational agents and machine translation among others. Traditionally, NLP tasks were achieved through rule-based systems, in essence, a set of manually crafted rules that determined the behaviour of the system. Examples include rule-based machine translation where linguists iteratively framed new rules to make the translations more accurate.

\par However, owing to the vast and heuristic nature of natural language, machine learning gained a stronger ground in performing NLP tasks. Machine learning models such as SVM, Naive Bayes and random forests found use in tasks such as sentiment analysis, spam detection and hate speech detection. On the other hand, Natural Language Generation(NLG) tasks such as machine translation, question-answering, abstractive summarization were achieved through models such as the Transformer and Seq2Seq architectures. 
\par Two important breakthroughs that provided significant impetus to the NLP and NLG domain were the arrival of Transfer Learning and rapid improvements in the performance of Language models. We\footnote[2]{Equal contribution of both the authors} feel it necessary to discuss these concepts before moving further:

\subsection{Language Modeling}Language modeling is an NLP task where the model has to predict the succeeding word given the previous words of the sequence as context. This task requires the language model(LM) to learn the nuances and inter-dependencies among the various words of the language. Standard benchmark datasets for the language modeling tasks include the wikitext dataset, the BookCorpus dataset\cite{bookcorpus} and the 1B word\cite{1bword}  benchmark. Perplexity is generally used as a metric to evaluate the performance of language models. Perplexity is defined as follows:
\begin{equation}
\sqrt[N]{\prod_{i=1}^{N}\frac{1}{P(w_i|w_1...w_{i-1})}}
\end{equation}
Given a sequence of $N$ words of the corpus, $w_1 w_2...w_N$, $P(w_i|w_1...w_{i-1})$ is the probability assigned by the language model to word $w_i$ given $w_{i-1}$ preceding words of the sequence.
\par A lower perplexity generally signals towards a better performing language model as it indicates a lower entropy in the generated text. Language modeling is used for tasks such as next word prediction, text auto-completion and checking linguistic acceptability. The concept of semi-supervised learning in NLP allows us to understand why and how language modeling has played a fundamental role in various architectures that have allowed for transfer learning in NLP.

\subsection{Transfer Learning}Traditionally, NLP models were trained after random initialization of the model parameters(also called weights). Transfer Learning, a technique where a neural network is fine-tuned on a specific task after being pre-trained on a general task allowed deep learning models to converge faster and with relatively lower requirements of fine-tuning data. Historically, transfer learning has been mainly associated with the fine-tuning of deep neural networks trained on the ImageNet dataset\cite{imgnet} for other computer vision tasks. However, with recent advances in natural language processing, it has become possible to perform transfer learning in this domain as well.
\par In this survey, we seek to discuss the recent strides made in Transfer learning, Language modeling and natural language generations through advancements in algorithms and techniques. Transfer learning can be used for applications where there is lack of a large training set. The target dataset should ideally be related to the pre-training dataset for effective transfer learning. This type of training is generally referred to as semi-supervised training where the neural network is first trained as a language model on a general dataset followed by supervised training on a labelled training dataset thus establishing a dependence of supervised fine-tuning on unsupervised language modeling.
\par The paper is structured as follows- Section II of the paper elaborates on the various algorithms and architectures that have serve as a base on which more advanced models have been built upon. Section III provides information about the transformer architecture which drastically improved the prospects of using transfer learning for NLP tasks. Section IV then goes on to discuss the evolution of language modeling and transfer learning through models such as BERT, ElMo, UlMFit and so on. We conclude our survey and suggest future improvements in section V.

\begin{figure}[h]
    
    \includegraphics[width=\linewidth]{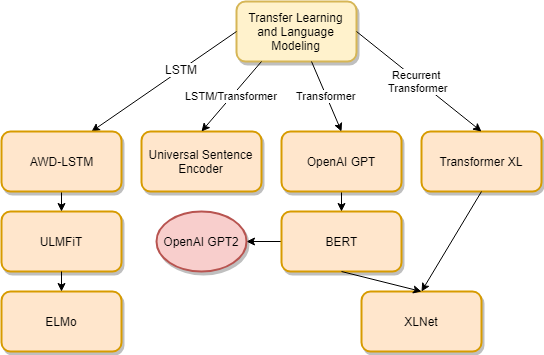}
    \caption{Developments in Transfer Learning and Language Modeling}
\end{figure}

\section{BACKGROUND}
\subsection{Vanilla RNNs}
Machine learning models have been widely been used for an array of supervised as well as unsupervised learning tasks such as regression, classification, clustering and recommendation modelling. Markov models such as the Multi-layer Perceptron Network, vector machines and logistic regression, however, did not perform well in sequence modeling tasks such as text classification, language modeling and tasks based on time series forecasting. These models suffered from an inability to retain information throughout the sequence and treated each input independently. In essence, the lack of a memory element precluded these models from performing well on sequence modelling tasks.\\
Recurrent Neural Networks\cite{rnn} or RNNs attempted to redress this shortcoming by introducing loops within the network, thus allowing the retention of information.

\begin{figure}[h]
    \centering
    \includegraphics[width=\linewidth]{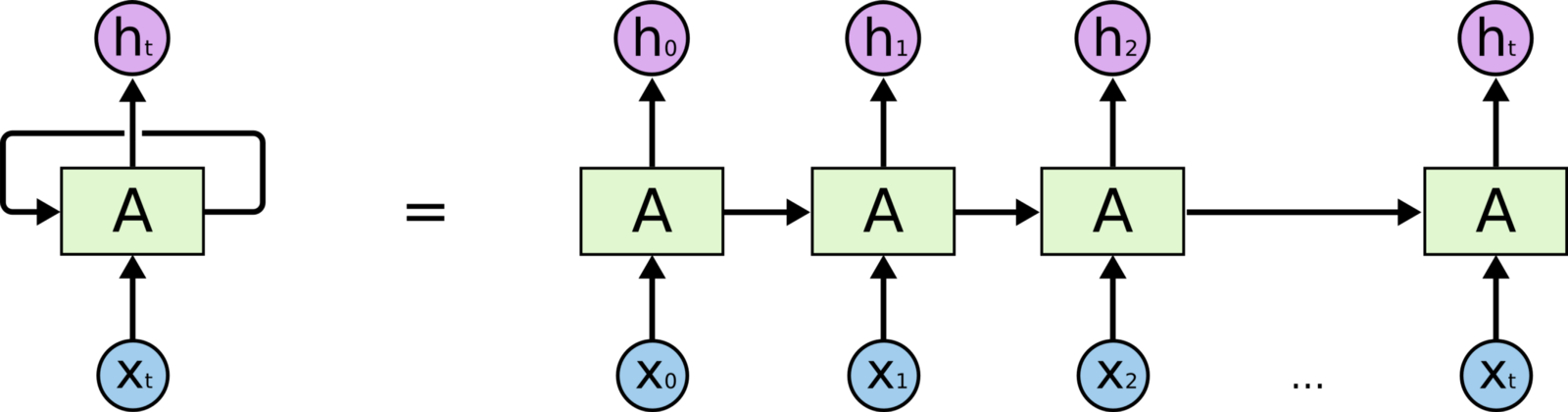}
    \caption{Recurrent Neural Network\cite{cola}}
\end{figure}
\begin{equation}
h_t = \phi(Wh_{t-1} + Ux_t + b)
\end{equation}

\par As shown in Fig.2 and the corresponding equation, the current hidden state of the neuron can be modelled as a function of the hidden state of the previous neuron $s_{t-1}$, the current input $x_t$, weight matrices \textit{ U, W} and bias \textit{b}.
These weights of the network are then updated through a training algorithm called Backpropagation Through Time(BPTT)\cite{bptime}.
\par BPTT is, in essence, the backpropagation algorithm with some modifications. The network is propagated for each time step- an operation that is often referred to as “unrolling” the RNN. The parameters of the neural network remain the same throughout the unrolling operation of the RNN. Corresponding errors concerning the predicted output and the ground truth are then calculated for each time step. Gradients of the error with respect to all parameters are then calculated and accumulated using the backpropagation algorithm. It is only after the unrolling is complete that all the parameters of the RNN are updated by using this “accumulated” error gradient.
\par Vanilla RNNs were successful in a wide variety of tasks including speech recognition, translation and language modelling. Despite their initial success, vanilla RNNs were only able to model short term dependencies. They failed to model long term dependencies, primarily because the information was often "forgotten" after the unit activations were multiplied several times by small numbers. Further, they suffered from various issues while training such as the vanishing gradient problem (the error gradient being used for weight updation reducing to very low values) and the exploding gradient problem. Thus, successfully training and applying vanilla RNNs was a challenging task. 

\subsection{Long Short Term Memory}
The problem of 'long-term' dependencies faced by earlier recurrent neural networks was solved by designing a special kind of RNN architecture called the LSTM(Long Short Term Memory)\cite{lstm}. They were designed to keep track of information for the extended number of timesteps. LSTMs have an overall chain-based architecture similar to RNNs but the crucial difference is the improvements in the internal node structure. While a node in RNN consists of a single neural layer, there are four layers connected uniquely in LSTMs as shown in the figure.
\begin{figure}[h]
    \centering
    \includegraphics[width=\linewidth,height=8cm]{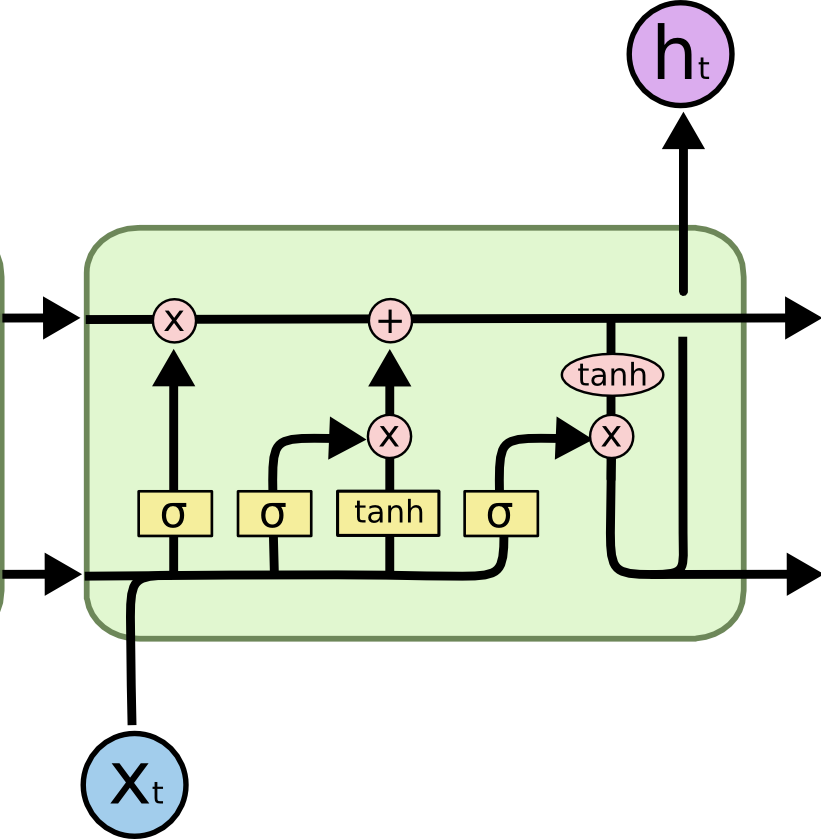}
    \caption{Chain structure of LSTMs\cite{cola}}

\end{figure}
\par The key feature of LSTMs is the information carrier connection present at the top known as the 'cell state'. It proves to be useful to carry information along longer distances with only minor linear operations taking place at each node.The ability to add or delete certain information of the cell state at each node is provided by structures called as gates. LSTM has three such gates, each comprising of a sigmoid neural net layer and a pointwise multiplier.
\par The 'forget gate' decides what information is to be retained in the cell state by using a sigmoid layer which outputs a value between 0 and 1(0 indicates 'forget everything' while 1 indicates 'retain completely'). The 'input gate' layer is used to determine new information which is to be added to the cell state. It involves deciding which values to be updated and the new candidates to do so. The previous cell state values and the new candidate values are then combined to get the final new cell state. The output to be forwarded by the node is decided by combining the values of current cell state with the results of the 'output layer'. The output is generally supporting information relevant to the previous word. Mathematically, the operations performed using the three gates can be expressed as:
 \begin{equation}
 f_t = \sigma(W_f.[h_{t-1},x_t] + b_f)
 \end{equation}
 \begin{equation}
 i_t = \sigma(W_i.[h_{t-1},x_t] + b_i)     
 \end{equation}
 \begin{equation}
 \widetilde{C}_t = tanh(W_c.[h_{t-1},x_t] + b_C)     
 \end{equation}
 \begin{equation}
 C_t = f_t*C_{t-1} + i_t*\widetilde{C}_t     
 \end{equation}
 \begin{equation}
 o_t = \sigma(W_o.[h_{t-1},x_t] + b_o)     
 \end{equation}
 \begin{equation}
 h_t = o_t*tanh(C_t)     
 \end{equation}

\par where $f_t$,$i_t$,$o_t$ indicate outputs of the forget gate, input gate and output gate respectively. \textit{W} and \textit{b} indicate the weights and bias. $C_{t-1}$ indicates the previous cell state, $\widetilde{C}_t$ represents the new candidate values and the current cell state is shown by $C_t$.
\par The advantage of using LSTM is that they offer more control in a network than the conventional recurrent networks. The system is more sophisticated and can retain information over longer timesteps. However, the added gates lead to more computation requirement and thus LSTMs tend to be slower.

\subsection{Gated Recurrent Units}
Gated Recurrent Units\cite{gru} or GRUs introduced by Cho, et al. in 2014 are a curtailed variation of LSTMs designed to reduce the computation issues of the latter. The forget and input gates in LSTMs are combined into a single 'update gate'. The cell state and hidden states are also merged together and computed using a single 'reset gate'. The operations now performed are as follows:
 \begin{equation}
 z_t = \sigma(W_z.[h_{t-1},x_t])     
 \end{equation}
 \begin{equation}
 r_t = \sigma(W_r.[h_{t-1},x_t])     
 \end{equation}
 \begin{equation}
 \widetilde{h}_t = tanh(W_c.[r_t*h_{t-1},x_t])     
 \end{equation}
 \begin{equation}
 h_t = (1-z_t)*h_{t-1} + z_t*\widetilde{h}_t     
 \end{equation}
 
\begin{figure}[h]
    \centering
    \includegraphics[width=\linewidth]{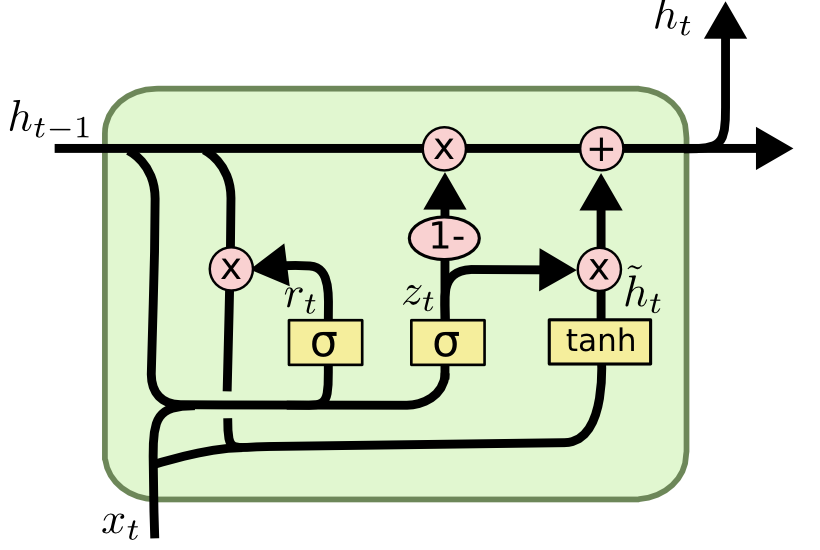}
    \caption{Internal structure of GRUs\cite{cola}}

\end{figure}
\par GRUs have the advantage of being able to control the flow of information without having an explicit memory unit, unlike LSTMs. It exposes the hidden content of the node without any control. The performance is almost on par with LSTM but with efficient computation. However, with large data LSTMs with higher expressiveness may lead to better results. 

\subsection{Average SGD Weight Dropped(AWD)-LSTM}
AWD-LSTM\cite{awd}, despite its relatively simple 3-layer LSTM architecture was proven to be highly effective for Language Modeling tasks. It employed a novel algorithm called DropConnect to mediate the problem of overfitting that had been inherent in the RNN architecture. Besides, the authors used Non-monotonically Triggered ASGD(NTASGD) algorithm to optimize the network.

\textbf{Dropconnect Algorithm} Neural networks, prone to overfitting, traditionally utilised Dropout as regularization to prevent overfitting. Dropout, an algorithm that randomly(with a probability p) ignore units' activations during the training phase allows for the regularization of a neural network. By diminishing the probability of neurons developing inter-dependencies, it increases the individual power of a neuron and thus reduces overfitting. However, dropout has not been able to provide commensurate results in case of the RNN architectures. In essence, it inhibits the RNN's capability of developing long term dependencies as there is loss of information caused due to randomly ignoring units activations.
\par To this end, the drop connect algorithm randomly drops weights instead of neuron activations. It does so by randomly(with probability 1-p) setting weights of the neural network to zero during the training phase. Thus redressing the issue of information loss in the Recurrent Neural Network while still performing regularization.

\textbf{ Non-monotonically Triggered ASGD (NT-ASGD)}
Stochastic gradient descent has been demonstrated to offer good performance for language modeling tasks through saddle point avoidance and linear convergence. Thus, the authors go on to investigate a variant of SGD- averaged SGD. Averaged SGD-almost identical to vanilla SGD- differs in the fact that an averaging of the weights(which are cached) is performed after a threshold number of iterations $T$ is over.
\par While theoretically able to control the effects of noise, averaged SGD has found little use while training neural networks. This has been mainly attributed by the author to ambiguous guidelines regarding the tuning of hyperparameters: learning rate scheduler and averaging trigger. A commonly used strategy while using the SGD optimizer is to reduce the learning rate by a fixed quantity when the validation error worsens or fails to improve. Similarly, one may perform the averaging operation after validation error worsens. The Non-monotonically triggered ASGD employs a similar technique. It differs in the fact that, instead of performing averaging when the validation error worsens NT-ASGD performs the averaging operation if the validation error fails to improve. NT-ASGD introduces two new hyperparameters-the logging interval L and the non-monotone interval n. Consequently, the authors found that keeping $n=5$ provided good performance in general. Better results were achieved compared to SGD while training their model. 

\subsection{Seq2Seq Architecture}
We take the example of neural machine translation to explain the working of the Attention Mechanism and advantages that it provides.
\par The Seq2Seq architecture\cite{seq2seq} has been used to perform a wide variety of tasks including Neural Machine Translation(NMT),  Abstractive summarization and chatbot systems. The traditional Seq2Seq architecture consists of an encoder RNN(LSTM/GRU) followed by a decoder RNN. The encoder encoded the given sequence into a fixed-length vector. The decoder generated the output sequence after taking the fixed-length vector as source hidden state. While giving significant improvements in the domains of neural machine translation, encoding the context of complex and long sequences into a single vector impeded the performance of the network. This was since a fixed-length vector was often incapable of effectively encoding the context of the given sequence. Consequently, this led to the birth of the Attention Mechanism, a novel technique that allowed the neural network to identify which input tokens are relevant to a corresponding target token in the output.

\subsection{Attention Mechanism}
 
Instead of encoding a single vector to represent the sequence, the attention mechanism\cite{attention} computes a context vector for all tokens in the input sequence for each token in the output. 
The decoder computes a relevancy score for all tokens on the input side. These scores are then normalized by performing a softmax operation to obtain the Attention weights. These weights are then used to perform a weighted sum of the encoder's hidden states, thus obtaining the Context Vector $c_t$.
\begin{equation}
\alpha_{ts} = \frac{exp(score(h_t,\overline{h}_s))}{\sum_{s'=1}^{S}exp(score(h_t,\overline{h}_{s'})))}    
\end{equation}
\begin{equation}
c_t = \sum_{s}\alpha_{ts}\overline{h}_s    
\end{equation}
\begin{equation}
a_t = f(c_t,h_t) = tanh(W_c[c_t;h_t])    
\end{equation}

A hyperbolic tangent operation is performed on the concatenation of the context vector and the target hidden state to get the Attention  vector $a_t$. This attention vector generally provides a better representation of the sequence than traditional fixed-sized vector methods.\newline
By identifying the relevant input tokens while generating the output token, Attention mechanism is able to redress the problem of compressing the context of the text into a fixed-sized vector.
Using this mechanism Bahdanu et. al.\cite{attention} were able to achieve state-of-the-art performance in machine translation tasks.

\begin{figure}[h]
    \centering
    \includegraphics[width=\linewidth]{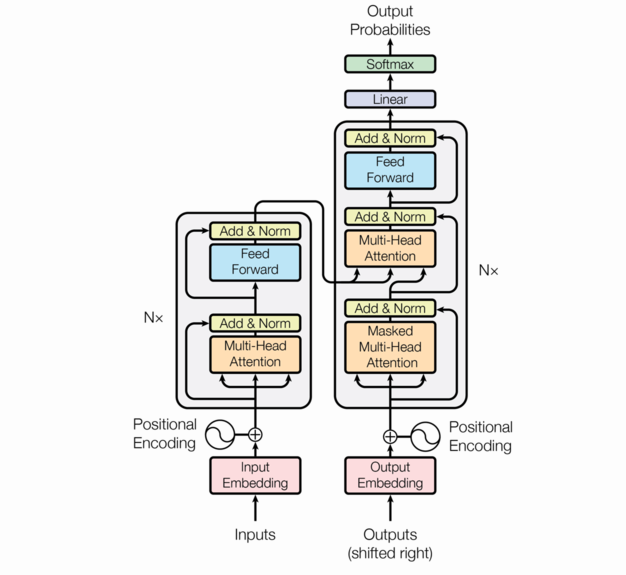}
    \caption{Architecture of the Transformer}
\end{figure}

\section{THE TRANSFORMER ARCHITECTURE}
Owing to the significant improvements gained due to the Attention mechanism, Vaswani et. al.\cite{Transformer} proposed the Transformer architecture. The Transformer achieved new state-of-the-art results in various tasks such as machine translation, entailment and so on. 
As shown in Fig.5, the Transformer consists of an encoder and a decoder. Furthermore, the encoder consists of a Multi-Head Attention layer, residual connections, normalization layer and a generic feed-forward Layer. The decoder is almost identical to the encoder but contains a certain "Masked" Multi-Head Attention layer.\newline \\
\textbf{Encoder: }The encoder takes as input the input embedding that is added with the positional encoding. The positional encoding allows for the retention of position and order related information. The authors employ the following equations to compute the positional encoding:
\begin{equation}
PE_{\left ( pos,2i \right )}=\sin (pos/10000^{\left ( 2i/d_{model} \right ) } )    
\end{equation}
\begin{equation}
PE_{\left ( pos,2i+1 \right )}=\cos (pos/10000^{\left ( 2i/d_{model} \right ) } )    
\end{equation}

 Where \textit{pos} is the position of the word in the sequence and \textit{i} is the dimension. This positional encoding is added to the input embedding. It is then followed by a residual connection \textit{R}, defined as follows:
 \begin{equation}
 R(x)= LayerNorm(x+MultiHeadedAttention(x))    
 \end{equation}
 Where \textit{x} is the value of the input embedding added with positional encoding. Thus we are effectively able to encode the semantic and position-related information using the input and positional encodings. \newline
\textbf{Decoder: } As previously stated, the decoder is almost identical to the Encoder but for the "Masked" Multi-Headed Attention Layer. \\
 \textbf{Scaled dot-Product Attention:} While generating embeddings for each word in the input token, the authors made use of the self attention mechanism. Self attention, similar to vanilla attention, allows the Transformer to identify words in the input sequence that were relevant to the current token.
Specifically, the authors made use of Scaled dot-Product Attention.
\begin{equation}
Attention\left(Q, K, V \right)=softmax( \frac{Q K^T}{\sqrt{d_k}} ) V     
\end{equation}

As shown above, the Scaled dot-Product Attention takes three vectors as input- the key, value and query.As shown, we perform a weighted average of the value vector $V$ . The weights are assigned by using a "compatibility function" to find . 
\begin{itemize}
    \item  Embedding $E$ is the output value of previous hidden layer
    \item  Query,  $Q=E W_q$ 
    \item  Key,  $Q=E W_k$ 
    \item  Value,  $Q=E W_v$ 
\end{itemize}
\par Where $W_q$, $W_k$ and $W_v$ are weight matrices.
A dot product of the query vector Q and key vector K is performed with a scaling factor $1/\sqrt{dk}$. The scaling factor is used to avoid the softmax input falling in a range where the output falls to a negligible value. A softmax operation is performed on the output of softmax. The final vector is multiplied with the value vector through a matrix multiplication operation to obtain the final attention scores. 
\begin{figure}[h]
    \centering
    \includegraphics[width=\linewidth]{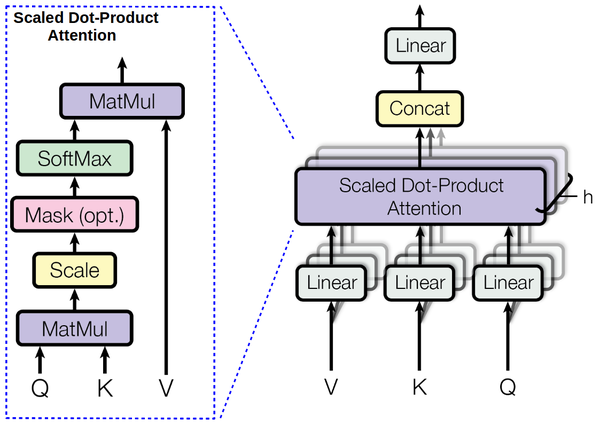}
    \caption{Multi-Headed and Scaled Dot-Product Attention}
\end{figure} \\
\textbf{Multi-Headed Attention:} Additionally, to reduce the number of operations to compute Attention scores, the authors make use of Multi-Headed Attention. Multi-Head Attention splits the vector space into 'n' parts. These divisions are then passed to 'n' Attention Heads to perform the Self-Attention operation, the results of these operations are then concatenated.
In addition to reducing the number of operations, Multi-Headed Attention allows the model to "jointly attend to information from different representation subspaces at different positions".

\textbf{ Masked Multi-Headed Attention:} The  Masked Multi-Headed Attention is similar to the Multi-Headed Attention but performs an additional "masking" operations. The decoder is allowed to attend to only the previous positions while computing self-attention passing the output embedding. This would result in the transformer being able to attend to the subsequent positions and consequently the output prediction in the sequence. To prevent this undesirable phenomenon, all subsequent positions are set to $-\infty $ before computing the self-attention that is passed to higher layers.\\ 
Finally, a softmax layer is used as the to compute output word probabilities. The word probabilities are in the form of a vector that has a size equal to the size of the vocabulary of the training set.
\par Based on the Attention mechanism and without using any recurrence mechanism, the Transformer effectively supplanted the Recurrent Neural Network Architectures-LSTM, GRU, etc.-as the state-of-the-art in several NLP tasks. It is used for a wide variety of tasks including machine translation and constituency parsing.

\section{EVOLUTION OF TRANSFER LEARNING AND LANGUAGE MODELING}
\subsection{ULMFIT}
Universal Language Model Fine-tuning (ULMFit)\cite{ulmfit}- a method to fine-tune a pre-trained language model- was one of the forerunners of inductive transfer learning in NLP. By definition, the language modeling task entails that given a sequence of tokens, the model has to predict the likelihood of the next token based on the sequence. The ULMFit method achieved good results by using the then state-of-the-art AWD-LSTM for their experiments. The proposed network was a simple 3-layer neural network without any attention mechanism, skip connections etc. ULMFit allowed for transfer learning by employing the following three steps in the stated order:

\begin{enumerate}
    \item {\textbf{Generic Pretraining of the Language Model}:
The authors pre-trained their language model on WikiText-103-- a large general-purpose dataset that consists of 28,595 preprocessed articles and 103 million words. This step allowed the language model to capture the general properties of the given language. Also, while computationally expensive, this task need to be performed only once.}
    \item {\textbf{Fine-tuning the Language Model on the Target task}:
This step allowed the language model to capture the inherent nuances of the target task,thus allowing better performance. Furthermore, this step can be performed on a relatively smaller dataset, thus requiring relatively less computational power. The authors then proposed two novel methods- Discriminative Fine-tuning and Slanted Triangular Learning rates to perform this task.\\
\textbf{Discriminative fine-tuning} - Different layers of the model extract different features from the text. Thus, the use of different learning rates for different layers seemed apt. To this end, the the authors formulated Discriminative Fine-Tuning, a variation of the SGD optimizer-as follows:
\begin{equation}
{\theta^l_t} = {\theta^l_{t-1}} - \eta^l\cdot \nabla_{\theta^l}J(\theta)     
\end{equation}

\par Where, $\theta$ is the weight, t is the iteration, l is the layer, $\nabla_{\theta^l}J(\theta)$ is the error gradient, $\eta$ is the learning rate and J indicates the error function.
 
 \par The authors found that choosing a value of learning rate for the last layers and then setting the learning rate of lower layers by the relation $\eta^{l-1} = \eta^{l}/2.6$ worked well.\newline

 \textbf{Slanted Triangular Learning Rates}: The Slanted Triangular Learning Rate is defined by the author as follows: by using a high initial learning rate, the model would quickly converge to an appropriate region in the hyperspace. The learning rate is then linearly decayed in order to improve the parameters on target task at a fine rate. The authors defined three new equations:\\ \\
  $cut = [T\cdot cut_{frac}] \\
  \\
p=\begin{Bmatrix} 
t/cut,  \;\;t<cut \\
\\ 1-\frac{t\ -\ cut}{cut\cdot (1/cut_{frac}\ -\ 1)},\ otherwise 
\end{Bmatrix} \\
\\ \\
\eta_t = \eta_{max} \cdot \frac{1\ +\ p\cdot (ratio\ -\ 1)}{ratio}$\\
    
    where $T$ is the number of iterations, $cut_{frac}$ the fraction of iterations that the LR is increased and \textit{p} is the number of iterations that the LR is going to be decreased or has been decreased. Additionally, \textit{ratio} defines $eta_{min}/eta_{max}$ and \textit{cut} is the cutoff iteration where the model switches from an increasing LR to a decreasing LR. }
    \item{\textbf{Fine-Tuning the Classifier on Target Task} }
    To perform task-specific classification, two linear blocks initialized from scratch are added to the language model. The author follows standard practices such as Batch Normalization and Dropout to perform regularization. Besides, the ReLu activation function is used similar to those used in Computer Vision models.\\
    \textbf{Concat Pooling}: To preserve information contained in few words, the input provided to this classifier is a concatenation of the last hidden layers and the average and max pooled output of the previous hidden layers. For this purpose, the author sought to concatenate as many hidden layers as would fit in the GPU memory. The concatenation $h_c$ is as given below:
     \begin{equation}
     h_c = concatenate(h_T, maxpool(H), averagepool(H))    
     \end{equation}
     
    $where\; H=\{h_1,h_2..h_T\}$ \newline\newline
    \textbf{Gradual Unfreezing}:
    Keeping all parameters trainable, i.e. performing the updation of all parameters during training would lead to a rapid loss in information learnt during the pre-training phase. To tackle this, the authors have gradually "unfrozen" the layers. In essence, the authors start by unfreezing the last layers and then perform fine-tuning. They repeat this process until all layers of the AWD-LSTM have not been trained. \\
    \textbf{Bidirectional LM}: The authors train both a forward LM as well as a backward LM. Consequently averaging the predictions given by both the Language Models.
\end{enumerate}

One can apply transfer learning using ULMFit by using pre-trained models trained on datasets such as the Wikitext 103 data.  Fine-tuning is performed by adding training the network on the target task by using supervised learning

\subsection{Embeddings from Language Models(ELMo)}
Traditional word embeddings involve assigning a unique vector to each word. These word embeddings are fixed and independent of the context in which the words are being used. Peters et. al came up with a new word representation called "Embeddings from Language Models(ELMo)",\cite{elmo} in which the tokens beings assigned to each word were a function of the entire sentence of which the word as a part of. These embeddings are obtained from the internal layers of a deep bidirectional LSTM that is trained with a coupled language model objective (biLM) on a large text corpus. These representations are more elaborate as they are dependent on all of the internal layers of the biLM. The word representations are computed on top of two-layer biLMs with character convolutions as a linear function of the internal network states. For a given set of tokens, the biLM computes their probability by taking into consideration the logarithmic likelihood of both the previous words(forward LM) as well as the future words(backward LM) and maximizing it.
 \begin{equation}
 \begin{split}
 p(t_1,t_2,...,t_N) = \sum_{k=1}^{N}(logp(t_k|t_1,...,t_{k-1};\Theta_x,\overrightarrow{\Theta}\textsubscript{LSTM},\Theta_s) + \\ logp(t_k|t_{k+1},...,t_N;\Theta_x,\overleftarrow{\Theta}\textsubscript{LSTM},\Theta_s))     
\end{split}
 \end{equation}
 
where $\Theta_s$ and $\Theta_x$ indicate the parameters for the softmax layer and the token representations in the forward and backward layer respectively. The ELMo model uses the intermediate layer representation of the biLM. ELMo combines all the layers of the biLM representation into a single vector $ELMO_k$ to be accommodated later in the fine-tuning task. Generally, for a given task of obtaining word embeddings in the language modeling phase, we find the obtained weightings of all biLM layers:
\begin{equation}
ELMo_k^{task} = \gamma^{task}\sum_{j=0}^{L}s_j^{task}h_{k,j}^{LM}
\end{equation}

where $s^{task}$ are the weights and $\gamma^{task}$ is a scalar quantity. $s^{task}$ is obtained after normalizing the weights and passing them through a softmax layer. $\gamma^{task}$, on the other hand, allows us to scale the ELMo vector. $\gamma^{task}$ plays an important role in the optimization process. The layer representation $h_{k,j}^{LM} = [\overrightarrow{h}^{\,LM}_{k,j};\overleftarrow{h}^{\,LM}_{k,j}]$ for each biLSTM layer wiz. combination of forward context representations and backward context representations.
\par Such a deep representation helps ELMo to trace two important factors (1) Complex characteristics like the syntax and semantics of the words being used and (2) Their variations across linguistic contexts. For any supervised task, the weights of the biLM are frozen. $ELMO_k$ is then concatenated with it and the obtained representations are then forwarded to the task-specific architecture. A pretrained biLM can be used to obtain representations for any tasks and with fine-tuning, they have shown a decrease in perplexity thereby benefiting from transfer learning. \\
The addition of ELMo embeddings to existing models has helped them process more useful information from the sentences and thus enhanced the performances in many applications like question answering, semantic role labelling, named entity extraction and many more. 
\par ELMo provides an enhancement to the traditional word embedding techniques by incorporating context while generating the same.  The vector generated by ELMo can be used for a general downstream task. This is sometimes done by passing the generated word embeddings to another neural network(eg. LSTM) which is then trained on the target task. Furthermore, concatenation of ELMo embeddings with other word embeddings is also done to provide better performance.
\subsection{OpenAI Transformer}
Although there is availability of large text corpora, labelled data is tough to find and manually labelling the data is an equally tedious task. Radford et al. at OpenAI proposed a semi-supervised approach called Generative Pre-training(GPT)\cite{gpt} which involved unsupervised pre-tuning of the model and then task-specific supervised fine-tuning for language understanding tasks. The Transformer is used as the base model for this purpose. The unsupervised learning helps to set the initial parameters of the model based on a language modeling objective. The subsequent supervised learning helps the parameters adjust to the target task.
\par Initially, a multi-layer transformer decoder is used to produce an output distribution over the target tokens based on a multi-headed self-attention mechanism.
\begin{equation}
h_0 = UW_c + W_p    
\end{equation}

\begin{equation}
h_l = transformer\_block(h_{l-1}) \forall i \in [1,n]    
\end{equation}

\begin{equation}
P(u) = softmax(h_nW^T_c)    
\end{equation}

where $h_i$ is the transformer layer's activations, $W_c$ is the token embedding matrix and $W_p$ is the position embedding matrix. The supervised learning task then obtains the final transformer block activations $h^l_m$ which are passed through a softmax layer to predict output label y:
\begin{equation}
P(y|x^1,...,x^m) = softmax(h^m_lW_y)    
\end{equation}
to maximize
\begin{equation}
L_2(C) = \sum_{(x,y)}logP(y|x^1,...,x^n)    
\end{equation}
where C indicates the labeled dataset with each sequence consisting of tokens $x^1,x^2,..x^m$. During transfer learning, the input is converted into a single contiguous sequence of tokens so as to fit the pre-trained model.
\par OpenAI transformer improvises on generative pre-training to improve performance on tasks by providing a better start than random initialization. A single model is able to produce quality results with minimum task-specific customization or hyperparameter tuning, thereby showing its robustness. This architecture model was able to outperform other approaches on tasks like natural language inference, question answering, sentence similarity etc.

\subsection{Bidirectional Encoder Represenation from Transformers(BERT)} BERT\cite{bert} -proposed by J.Devlin et al.- is a novel approach to incorporate bidirectionality in a single Transformer model. A particularly challenging task, direct approaches to incorporating bidirectionality in Transformer models fail since direct bidirectional conditioning would allow the words to see themselves in the light of context from multiple layers, thereby ruling out the possibility of using it as a Language Model. In essence, it was traditionally only possible to train a unidirectional encoder- a left-right or a right-left model. However, bidirectional models that could see the complete sequence context would inherently be more powerful than unidirectional models or a concatenation of two unidirectional models-left-right and right-left. To this end, the authors trained their model on two unsupervised prediction tasks:
\par \textbf{Masked LM} To overcome the challenges posed while applying of bi-directionality in Transformers, J.Devlin proposed masking of random tokens in the sequence. The Transformer was trained such that it had to predict only the words that had been masked while being able to view the whole sequence. WordPiece Tokenization is used to generate the sequence of tokens where rare words are split into sub-tokens. Masking of 15\% of the Wordpiece Tokens is performed. Masking essentially replace the words with $[MASK]$ tokens. However, instead of always replacing the selected words with a $[MASK]$ token, the data generator employs the following approach:
\begin{itemize}
\item Replace the word with $[MASK]$ token 80\% of the time
\item Replace the word with another random word 10\% of time
\item Keep the word as it is 10\% of the time
\end{itemize}
Performing prediction on only 15\% of all words instead of performing prediction on all words would entail that BERT would be much slower to converge. However, BERT showed immediate improvements in absolute accuracy while converging at a slightly slower pace than traditional unidirectional left-right models.
\par \textbf{Next Sentence Prediction} This task entails predicting whether the first sequence provided immediately precedes the next. This task allows the Transformer to perform better on several downstream tasks such as question-answering, Natural Language Inference that involve understanding the relationship between two input sequences. The dataset so used for training had a balanced 50/50 distribution created as follows: choosing an actual pair of neighbouring sentences for positive examples and a random choice of the second sentence for the negative examples. The input sequence for this pair classification task is generated as:\newline
$[CLS] <Sentence\; A> [SEP] <Sentence\; B> [SEP]$,
\par where sentences A and B are two sentences
after performing the masking operations. The [CLS] token is the first token used to obtain a fixed vector representation that is consequently used for classification and [SEP] is used to separate the two input sequence.
The authors were able to achieve an impressive accuracy of 97-98\% in the next sentence prediction task. 
\par \textbf{Pre-Training Procedure} The authors have used the BooksCorpus and the English Wikipedia as pretraining data. They have used two variations of BERT- $BERT_{BASE}$(12-layer) and $BERT_{LARGE}$(24 layers)- that primarily differ in their depth. The maximum length of the input sequence is restricted to 512 tokens. All subsequent tokens in the sequence are neglected. A dropout value of 0.1 is used as regularization. Furthermore, the authors have made use of the GELU instead of Relu as activation function. GELU- Gaussian Error Linear units has been shown to provide improvements compared to ReLU and eLu. 
\par Training of the models was performed on TPUs, specifically $BERT_{BASE}$ was trained on 16 TPU chips for 4 days. $BERT_{LARGE}$ was trained on 64 TPU chips, also for 4 days.
\par \textbf{Fine-Tuning Procedure} The pre-trained BERT can be fine-tuned on a relatively small dataset and requires lesser processing power.  BERT was able to improve upon the previous state-of-the-art in several tasks involving natural language inference, question answering, semantic similarity, linguistic acceptability among other tasks. The pattern of the input and output sequence varies depending on the type of the task. The tasks can be broadly divided into four categories:
\begin{itemize}
    \item Single Sentence Classification Tasks:
    These tasks are performed by adding layers on the classification embedding $[CLS]$ and passing the input sequence preceded by the $[CLS]$ token.
    \item Sentence Pair Classification Tasks
    The two sentences are passed to BERT after being separated by the $[SEP]$ token. Classification can be performed by adding layers to the $[CLS$
    \item Question Answering Tasks
    \item Single Sentence Tagging Tasks
\end{itemize}
Subsequently, two multilingual BERT models-uncased and cased-for over 102 languages were released.
Furthermore, OpenAI released the GPT2\cite{gpt2}, essentially BERT trained as a language model on a very large amount of data.
\subsection{Universal sentence encoder}
The amount of supervised training data present for language processing tasks is limited. The use of pre-trained word embeddings has proved to be useful in this case as they perform a limited amount of transfer learning. Daniel et al. \cite{universalSE} proposed a new approach which involved direct encoding of sentences instead of words into vectors. The sentence encoded vectors are found to require minimal task-specific data to produce good results. The encoder models are available in two architectures taking into consideration the two primal challenges of training transfer learning models wiz. complexity and accuracy.
\subsubsection{Transformer based architecture}
The first model makes use of the transformer architecture to construct sentence embeddings. The encoding subgraph of the transformer architecture is used for this purpose. Attention mechanism is used to find context-based word representations which are then converted into a fixed-length sentence encoding vector. The input to the transformer model is a lower case Penn Treebank 3(PTB)\cite{ptb} tokenized string and a 512-dimensional sentence embedding is produced as the output. A single encoding model is trained over multiple tasks to make it as general as possible. This model achieves superior accuracy over the other architecture but at the expense of increased computation requirement and complexity.
\subsubsection{Deep Averaging Network(DAN) architecture} 
In this model, the input embeddings for words and bi-grams are averaged and then passed through a deep neural network. The input and output format are same as that of the transformer encoder. Multitask learning, similar to the transformer model encoder model is used for training purpose. The main advantage of this model is that it performs the required operations in linear time.
\par The main difference between the transformer and DAN encoder models is of time complexity(O($n^2$) and O(n) respectively). The memory requirement for the transformer model increases quickly with increase in sentence length while that of DAN model remains constant. The trade-off between complexity and accuracy should be noted when deciding a particular architecture for a given task.
\par The unsupervised learning data used for training in both the cases included Wikipedia, web news and discussion forums. Augmentation is performed using training on supervised data from the Stanford Natural Language Inference(SNLI) corpus which improved the performance further. The universal sentence encoder can be used for a variety of transfer tasks including sentiment analysis, sentence classification, text similarity etc. For determining a pairwise semantic similarity between two sentences, the similarity of the sentence embeddings produced by the encoder can be calculated and converted into angular distance to get the final result.
\begin{equation}
sim(u,v) = (1 - arccos(\frac{u.v}{||u|| ||v||})/\pi)    
\end{equation}

The sentence embeddings outperform the results of word embeddings on the fore mentioned tasks. However, combining word and sentence embeddings for transfer learning produced the best overall results. Universal sentence encoder assists the most when limited training data is available for the transfer task. 
\par The Universal Sentence Encoder can be used for downstream tasks bypassing the generated embedding to a classifier such as an SVM or another deep neural network.
\subsection{Transformer-XL}
The Transformer-XL\cite{txl}was able to model very long-range dependencies. It did so by overcoming one limitation of the vanilla Transformer- fixed-length context. Vanilla Transformers were incapable of accommodating a very long sequence owing to this limitation. Hence they resorted to alternatives such as splitting the corpus into segments which could be managed by the Transformers. This led to loss of context among individual segments despite being part of the same corpus. However, the Transformer-XL was able to take the entire large corpus as input, thus preserving this contextual information. In essence a vanilla Transformer, it relied on two novel techniques-Recurrence mechanism and Positional Encoding- to provide the improvement.
\subsubsection{Recurrence mechanism}Instead of training the Transformer on individual segments of the corpus(without regards to previous context), the authors propose caching the hidden sequence state computed from the previous segments. Consequently, the model computes self attention(and other operations) on the current hidden/input state as well as these cached hidden states. The number of states cached during training is limited due to memory limitations of the GPU. However, during inference, the authors can increase the number of cached hidden states used to model long-term dependency. 
\subsubsection{Relative Positional Encoding}An inherent problem with using the said Recurrence mechanism is preserving relative positional information while reusing cached states. The authors overcame this problem by incorporating Relative Positional Encoding in the Attention mechanism(instead of hidden states) of the Transformer. They do so by encoding information regarding the positional embedding dynamically in the Attention scores themselves. The distance of the key vectors for the query vector is the temporal information that is encoded in the Attention scores. In essence, computing attention is facilitated as temporal distance is still available to the model while still preserving previous states. Furthermore, information regarding the absolute position can be recovered recursively.
\par The Transformer-XL was able to surpass the state-of-the-art results for language modelling tasks on the \textit{enwiki8} and \textit{text8} datasets.

\subsection{XLNet}
The BERT model proposed by Authors et. al, was an Auto Encoding(AE) model that suffered from the following problems:
\begin{itemize}
    \item The use of $[MASK]$ tokens during the pre-training phase led to a discrepancy as these tokens were absent during the fine-tuning phase.
    \item The model neglected inherent dependencies among two or more $[MASK]$ tokens, thus leading to sub-optimal performance.
\end{itemize}
A new model, the XLNet\cite{xlnet} was able to overcome these difficulties by using a modification of general autoregressive pretraining. 
\par \textbf{Generalized Autoregressive Pretraining Phase} Instead of using unidirectional language modeling or bidirectional masked language model to predict tokens, the paper proposed passing all permutations of a given sequence to the model and predicting a particular token missing from this sequence. Despite random re-ordering of the sequence, order-related information remains preserved as positional encodings of the tokens remained the same for all permutations of the input sequence.
\par Use of this modified form of pretraining helped overcome the two main challenges posed by the BERT architecture. Along with that, the XLNet incorporated Transformer-XL into its core architecture. This allowed for better modeling of long-dependencies compared to BERT. Through the use of these two major modifications, the XLNet provided new state-of-the-art results in 18 natural language processing tasks.
\par Significant gains were observed compared to BERT especially in tasks such as machine reading comprehension which required modeling of long-range dependencies. The authors attribute this improvement mainly to the use of Transformer-XL in the XLNet architecture. The XLNet, similar to BERT, can be used for a wide range of single sentence, sentence pair and reading comprehension tasks among others.

\section{CONCLUSION}
We have thus provided a lucid summary of recent advances in the domain of transfer learning in the domain of natural language processing. We hope that this survey would help the reader gain a quick and profound understanding of this domain. Recent advances in this domain, despite being a step forward, come with their challenges. Specifically, large architectures such as the BERT, XLNet and Transformer-XL make training and deployment difficult owing to the large amount of processing power required. Furthermore, employing large and opaque models impedes upon the explainability aspect of the same, thus making one question their deployment in the real-world. Thirdly, while newer models can provide improvements over their predecessors, the lack of a standard benchmark dataset for pre-training these models makes one question whether these improvements were due to an architectural innovation or simply because the said model was pre-trained on larger amount of data. Take for instance, it is difficult to gauge whether the XLNet model bettered upon the BERT model because of an architectural improvement or because it was pre-trained on a larger corpus. Thus, there is a need to decide upon a standard pre-training dataset to remove this ambiguity. Lighter models such as the DistilBERT and ALBERT are a step in the right direction and could potentially help bridge the gap between performance and processing power. On the other hand, innovations brought about during the training phase, such as in the RoBERTa model might help seek out better performance using the same model architecture.
\addtolength{\textheight}{-12cm}   

\end{document}